\documentclass{article}

\usepackage{arxiv}
\usepackage[utf8]{inputenc} % allow utf-8 input
\usepackage[T1]{fontenc}    % use 8-bit T1 fonts
\usepackage{hyperref}       % hyperlinks
\usepackage{url}            % simple URL typesetting
\usepackage{booktabs}       % professional-quality tables
\usepackage{amsfonts}       % blackboard math symbols
\usepackage{nicefrac}       % compact symbols for 1/2, etc.
\usepackage{microtype}      % microtypography
\usepackage{graphicx}
\usepackage{subcaption}
\usepackage{tabularx}
\usepackage{multirow}
\usepackage{amssymb}
\usepackage{wrapfig}
\usepackage{pgfplots}
\usepackage{tikz}
\usepackage{fancyvrb}
\pgfplotsset{compat=1.18}
\usepackage{tcolorbox}
\tcbuselibrary{breakable}
\usepackage{makecell}
\usepackage{natbib}
\usepackage{amsmath}
\usepackage{listings}
\usepackage{xcolor}

\definecolor{delim}{RGB}{20,105,176}
\definecolor{numb}{RGB}{106, 109, 32}
\definecolor{string}{rgb}{0.64,0.08,0.08}

\lstdefinelanguage{json}{
    numbers=left,
    numberstyle=\small,
    frame=single,
    rulecolor=\color{black},
    showspaces=false,
    showtabs=false,
    breaklines=true,
    postbreak=\raisebox{0ex}[0ex][0ex]{\ensuremath{\color{gray}\hookrightarrow\space}},
    breakatwhitespace=true,
    basicstyle=\ttfamily\small,
    upquote=true,
    morestring=[b]",
    stringstyle=\color{string},
    literate=
     *{0}{{{\color{numb}0}}}{1}
      {1}{{{\color{numb}1}}}{1}
      {2}{{{\color{numb}2}}}{1}
      {3}{{{\color{numb}3}}}{1}
      {4}{{{\color{numb}4}}}{1}
      {5}{{{\color{numb}5}}}{1}
      {6}{{{\color{numb}6}}}{1}
      {7}{{{\color{numb}7}}}{1}
      {8}{{{\color{numb}8}}}{1}
      {9}{{{\color{numb}9}}}{1}
      {\{}{{{\color{delim}{\{}}}}{1}
      {\}}{{{\color{delim}{\}}}}}{1}
      {[}{{{\color{delim}{[}}}}{1}
      {]}{{{\color{delim}{]}}}}{1},
}

\title{The Emergence of Altruism in Large-Language-Model Agents Society}

\author{
 Haoyang Li \\
  Department of Sociology\\
  Hong Kong Baptist University\\
  Hong Kong, China \\
  \texttt{LI\_Haoyang@life.hkbu.edu.hk} \\
 \And
 Xiao Jia \\
  School of Artificial Intelligence\\
  The Chinese University of Hong Kong, Shenzhen\\
  Shenzhen, China \\
  \texttt{225080011@link.cuhk.edu.cn} \\
 \And
 Zhanzhan Zhao \thanks{Corresponding author.}\\
  School of Artificial Intelligence\\
  School of Humanities and Social Science\\
  The Chinese University of Hong Kong, Shenzhen\\
  Shenzhen, China \\
  \texttt{Zhanzhanzhao@cuhk.edu.cn} \\
}

\begin{document}
\maketitle
\begin{abstract}
Leveraging Large Language Models (LLMs) for social simulation is a frontier in computational social science. Understanding the social logics these agents embody is critical to this attempt. However, existing research has primarily focused on cooperation in small-scale, task-oriented games, overlooking how altruism, which means sacrificing self-interest for collective benefit, emerges in large-scale agent societies. To address this gap, we introduce a Schelling-variant urban migration model that creates a social dilemma, compelling over 200 LLM agents to navigate an explicit conflict between egoistic (personal utility) and altruistic (system utility) goals. Our central finding is a fundamental difference in the social tendencies of LLMs. We identify two distinct archetypes: "Adaptive Egoists" (e.g., \texttt{o1-mini}, \texttt{o3-mini}), which default to prioritizing self-interest but but whose altruistic behaviors significantly increase under the influence of a social norm-setting message board; and "Altruistic Optimizers" (e.g., \texttt{Gemini-2.5-pro}, \texttt{Deepseek-R1}), which exhibit an inherent altruistic logic, consistently prioritizing collective benefit even at a direct cost to themselves. Furthermore, to qualitatively analyze the cognitive underpinnings of these decisions, we introduce a method inspired by Grounded Theory to systematically code agent reasoning. In summary, this research provides the first evidence of intrinsic heterogeneity in the egoistic and altruistic tendencies of different LLMs. We propose that for social simulation, model selection is not merely a matter of choosing reasoning capability, but of choosing an intrinsic social action logic. While "Adaptive Egoists" may offer a more suitable choice for simulating complex human societies, "Altruistic Optimizers" are better suited for modeling idealized pro-social actors or scenarios where collective welfare is the primary consideration.
\end{abstract}

\section{Introduction}
The application of Large Language Models (LLMs) has emerged as a promising research paradigm within computational social science and related fields \citep{anthis2025llmsocialsimulationspromising, tang-etal-2025-gensim}, such as such as urban computing and population modeling \citep{yan2024opencityscalableplatformsimulate, karten2025llmeconomistlargepopulation}. One of principal applications in this area is LLM-driven agent-based modeling (ABM), which utilize LLMs to simulate complex social environments and predict the behavior of LLM-driven agents therein \citep{10.1145/3586183.3606763, gao-2024}. Exhibiting anthropomorphic characteristics and outstanding role-playing capabilities, LLM-driven agents address critical limitations of traditional modeling approaches \citep{taillandier2025integratingllmagentbasedsocial, haase2025staticresponsesmultiagentllm}. Specifically, they transcend the rigid, predefined rules inherent in traditional methods. Furthermore, by articulating their reasoning processes, LLM-driven ABM can overcome the interpretability challenges of older methods \citep{gao2023largelanguagemodelsempowered, 10.5555/3600270.3602070}.

\begin{figure}[h]
  \centering
  \includegraphics[width=\textwidth]{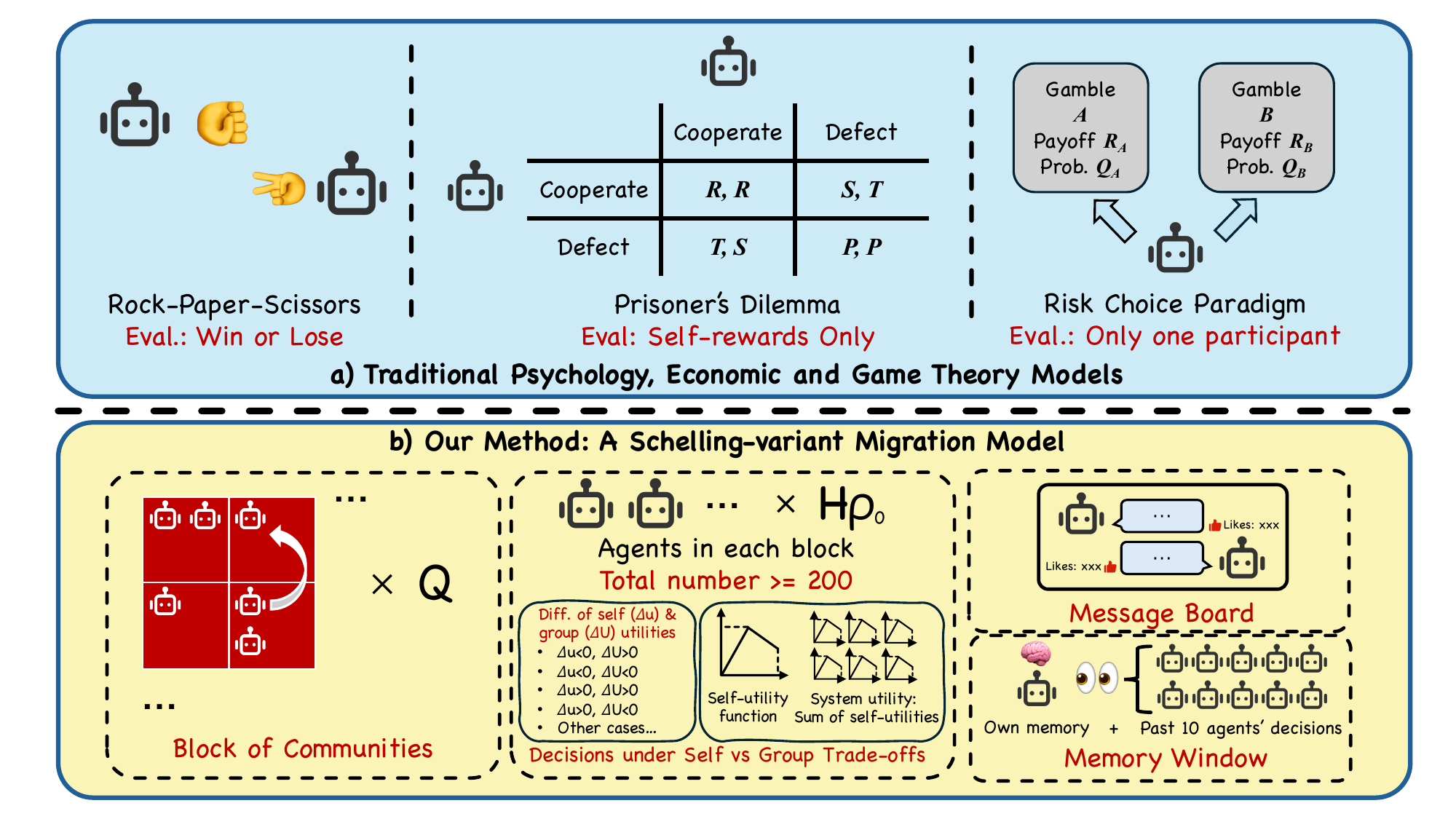}
  \caption{Through a Schelling-variant migration model, our method can capture the emergence of diverse rationalities within a complex social environment.}
  \label{fig:pic1}
\end{figure}

A primary application of LLM-driven ABM is social simulation, a field predicated on the human-like reasoning and behavioral capabilities of LLMs, whose formidable reasoning abilities has been validated across mutiple benchmarks \citep{haase2025staticresponsesmultiagentllm, deepseekai2025deepseekr1incentivizingreasoningcapability, openai2024gpt4technicalreport, comanici2025gemini25pushingfrontier, cobbe2021gsm8k}. However, the existing literature on LLM-driven social simulation frequently overlooks a fundamental aspect of these models' intrinsic nature when they engage in group-dynamic decision-making: their disposition towards egoistic versus altruistic choices, particularly in scenarios where individual and collective interests diverge. This oversight necessitates a critical distinction: while much of the current work in LLM-driven social simulation aims to simulate humans by embedding agents with detailed personas to replicate real-world behavior \citep{yan2024opencityscalableplatformsimulate, karten2025llmeconomistlargepopulation, piao2025agentsocietylargescalesimulationllmdriven}, this study focuses on a different yet more foundational goal of understanding the "LLM Society" itself. We treat LLMs as a new class of intelligent agents, placing them in a well-designed environment to observe their intrinsic, spontaneously formed social structures, norms, and behaviors. We posit that a deep understanding of an LLM's inherent natur, for instance, whether it is an innate utilitarian or a socially influenced egoist, is a prerequisite for examing whether it is suitable for using as a key tools for social simulation.

As stated, the using of LLM for social simulation fundamentally rely on an LLM's capacity for human-like social decision-making. Debates on human rationality in the decision-making process have long been central to social science. Classical economics posits the \textit{homo economicus}, a model of perfect rationality where agents act with pure self-interest to maximize personal utility \citep{tittenbrun-2013, mele-2014, 10.1257/jep.9.2.221}. In contrast, theories of bounded rationality argue that humans, constrained by cognitive and environmental limits, make "satisficing" rather than optimal choices \citep{Simon1957, Simon1976}. This gives rise to the tension between egoism, which prioritizes individual gain, and altruism, which serves the group's interests. An egoistic choice might align with perfect rationality at the individual level, yet altruistic act, which may appear irrational from a purely selfish viewpoint, are crucial for collective success. It is vital to distinguish this true altruism from "cooperation" often seen in game theory, which can be a long-term egoistic strategy \citep{willis2025systemsllmagentscooperate}. We focus on altruism as an immediate sacrifice of personal utility for collective gain.

Given their role as a key tool for social simulation, LLMs must be evaluated in contexts reflecting these complex dynamics. However, current evaluation benchmarks are often confined to micro-level scenarios from game theory, such as the Iterated Prisoner's Dilemma (IPD), Rock-Paper-Scissors, or resource allocation games \citep{Fontana_Pierri_Aiello_2025, zheng2025nashequilibriumboundedrationality, mao-etal-2025-alympics, wang2024battleagentbenchbenchmarkevaluatingcooperation, 10.5555/3709347.3744012}, where a small number of agents make decisions in direct engagements without further settings of social environment and interactions with the group to probe their rationality or strategic tendencies (e.g. egoism and altruism). This approach fails to capture the emergent phenomena of real-world situations. At the macro-level of society consisting of a large number of agents, decision-making is not always about direct, zero-sum competition but about navigating the trade-offs between individual and group welfare without centralized coordination.

To address this gap, we propose a novel Schelling-variant model of urban migration \citep{Schelling-1969, doi:10.1073/pnas.0906263106}. As a classic model, Schelling's Model demonstrates the universal mechanisms of residential migration and segregation through simple and verifiable interaction rules. It reveals broadly applicable social patterns without needing details for any specific community.  As shown in Figure \ref{fig:pic1}, our framework is designed for large-scale social simulation, creating a social dilemma where over 200 agents must weigh personal utility against group goals. To simulate social influence, we introduce a shared message board, a mechanism designed to foster pro-social norms.  We evaluate the emergent behaviors using a mixed-methods approach. Quantitatively, we assess macro-level system states using metrics for collective welfare and population distribution, while also analyzing micro-level dynamics by classifying each agent's action based on its simultaneous impact on individual and system utility. To complement this quantitative analysis and decode the reasoning behind agent choices, we also pioneer a qualitative analysis using an LLM-as-judge, inspired by Grounded Theory. This comprehensive setup allows us to investigate the following research questions:

\begin{itemize}
    \item \textbf{RQ1:} In a society with conflicting interests, do LLM agents default to egoism, or can they spontaneously generate altruistic behavior?
    \item \textbf{RQ2:} How does a social communication mechanism influence the balance between egoistic and altruistic behavior?
    \item \textbf{RQ3:} Do different LLM models exhibit fundamental, stable differences in their egoistic versus altruistic inclinations?
\end{itemize}

Our experiments reveal a fundamental bifurcation in the social tendencies of LLMs. We identify a class of "Adaptive Egoists" (e.g., \texttt{o1-mini}, \texttt{o3-mini}, \texttt{Qwen2.5-7B}), which default to self-interest but demonstrate significantly increased altruism when exposed to social norms via the message board. In contrast, we find a class of "Altruistic Optimizers" (e.g., \texttt{Gemini-2.5-pro}, \texttt{Deepseek} Series), which consistently pursue the collective good, displaying an inherent altruistic logic that leads them to prioritize system-level optima, even at a personal cost. These findings suggest that for social simulation, the choice of an LLM constitutes a choice of the simulation's underlying theoretical foundation. We argue that the suitability of an LLM agent for social simulation is context-dependent. The nuanced, socially-influenced behavior of "Adaptive Egoists" offers a higher-fidelity tools for simulating the complex, often sub-optimal bounded rationality dynamics of human societies. Relatively, "Altruistic Optimizers" are better suited for modeling idealized pro-social actors, such as in theories of collective action or resource optimization for the common good.

\section{Related Work}

\subsection{LLM Agents in Large-Scale Social Simulation}
The application of LLMs to social simulation has become a prominent research direction, leveraging their capabilities to create increasingly sophisticated simulations \citep{10.1145/3586183.3606763, anthis2025llmsocialsimulationspromising}. Foundational work has established that LLMs excel in complex reasoning tasks such as mathematics, coding, and planning \citep{openai2024gpt4technicalreport, chen2021codex, cobbe2021gsm8k}. Building on this, numerous studies have successfully employed LLMs to simulate intricate social phenomena, including urban activities \citep{yan2024opencityscalableplatformsimulate}, mobility \citep{wang2024large}, economic behaviors \citep{li-etal-2024-econagent, karten2025llmeconomistlargepopulation}, crime predictions \citep{zeng2025crimemindsimulatingurbancrime}, and even large-scale societal simulation with over 100,000 agents such as AgentSociety \citep{piao2025agentsocietylargescalesimulationllmdriven}, SocioVerse \citep{zhang2025socioverseworldmodelsocial}, GenSim \citep{tang-etal-2025-gensim}, and YuLan-OneSim \citep{wang2025yulanonesimgenerationsocialsimulator}.

Moreover, a significant methodological challenge in current LLM-driven social simulations is the tendency of agents to exhibit an average persona, lacking the behavioral heterogeneity crucial for realistic social dynamics \citep{anthis2025llmsocialsimulationspromising, wu2025llmbasedsocialsimulationsrequire}. Researchers have noted that because LLMs are trained to capture mainstream patterns, they often converge on uniform behaviors, which can kill the various characteristics and diverse strategies that drive complex social outcomes in the real world \citep{wu2025llmbasedsocialsimulationsrequire}. This limitation raises a critical question that precedes the goal of replicating human patterns: what are the fundamental social tendencies of different LLMs themselves?

Our research addresses this fundamental issue concerning the micro-foundations of these large-scale agent societies by drawing upon established sociological frameworks that explain the causal pathways between individual actions and societal outcomes. Our explanatory logic is guided by James Coleman's foundational model of Coleman's Boat \citep{coleman-1990}, which details how macro-level social conditions influence the micro-level behavior of individuals, which in turn aggregates to new macro-level results. Methodologically, our simulation builds on the pioneering work of Thomas Schelling \citep{schelling-2006}, who demonstrated how complex macro-level patterns, such as social segregation, can emerge from simple, individual-level motives. By integrating these frameworks, we investigate a core question: when individual and collective interests diverge, as structured by our simulation environment, what are the intrinsic social tendencies of LLM agents? Do they default to egoism, or can they manifest altruism?

\subsection{From Cooperation in Games to Altruism in Societies}

Recent studies have begun to probe the social behaviors of LLMs using classic paradigms from game theory psychology, and economics. These works have provided valuable insights by employing games like the IPD \citep{Fontana_Pierri_Aiello_2025, willis2025systemsllmagentscooperate}, Rock-Paper-Scissors \citep{zheng2025nashequilibriumboundedrationality}, Texas Hold'em \cite{guo2024suspicionagentplayingimperfectinformation}, Werewolf, bargaining, and coding tasks \citep{zhu2025multiagentbenchevaluatingcollaborationcompetition}, other classic games from game theory \citep{huang2025competing}, and resource allocation scenarios \citep{mao-etal-2025-alympics}. For evaluating decision-making patterns, studies have also adopted risk preference paradigms from psychology, using gambles with varying probabilities and payoffs to measure alignment with human decision-making \citep{liu2025large}.

However, these studies face two fundamental limitations. First, they are typically confined to dyadic or small-group interactions, which cannot capture the emergent group dynamics and social norm formation present in large-scale societies. Second, they primarily measure cooperation. Their experimental setups often involve too few agents to constitute a society. This scale is insufficient to meaningfully examine how agents weigh collective interests, let alone to construct the complex decision-making scenarios where individual and collective interests conflict. Addressing this gap, our experiment is designed to investigate a stronger form of altruism: the willingness to incur personal costs for the collective good. This concept is closely aligned with "strong reciprocity" from behavioral economics, wherein individuals voluntarily punish norm violators at a personal cost to uphold social cooperation, without the expectation of material reward \citep{trivers-1971, gintis-2000}. Additionally, their evaluation metrics are often based on simple win-or-loss outcomes or focus solely on human-agent alignment, lacking a horizontal comparison across different LLMs. Our research framework, based on a Schelling-variant model, addresses these gaps by supporting complex group decision-making with over 200 agents. It facilitates a comparative analysis of different LLMs' performance and alturism under various conditions.

\section{Problem Formulation}

We formulate the Schelling-variant urban residential migration model as a multi-agent system. The environment consists of a set of residential community blocks, denoted by $\mathcal{Q}$. Each block $q \in \mathcal{Q}$ has an identical maximum carrying capacity of $H$. The simulation is initialized with a global population density of $\rho_0$, resulting in a total of $N = \mathcal{Q} \cdot H \cdot \rho_0$ LLM-driven resident agents. At initialization ($t=0$), these $N$ agents are randomly assigned to the blocks in $\mathcal{Q}$, creating an initial location distribution $L_0$ with a degree of randomness in block-level densities. The simulation then begin with discrete timesteps $t=1, 2, \dots, T$.

\subsection{System Dynamics: State, Action, and Transition}

The global state of the environment at timestep $t$ is $s_t \in \mathcal{S}$, defined by the tuple $s_t = (L_t, \mathcal{H}_t, B_t)$. $L_t$ is the vector of all agent locations, $\mathcal{H}_t$ is the complete history recording all past actions and their corresponding system states, and $B_t$ is the state of the public message board, which contains a list of anonymous messages and their associated like counts.

\sloppy{Based on its observation (defined in section \ref{Agent Observation and Objectives}), each agent $i$ takes a composite action $a_i^t \in \mathcal{A}_i$, represented by a tuple $a_i^t = (a_{\text{move}}^i, a_{\text{board}}^i)$, where $a_{\text{move}}^i$ is the migration decision (\texttt{stay} or \texttt{move\_to(q)}) and $a_{\text{board}}^i$ is the message board action (\texttt{post(message)}, \texttt{like(message\_id)}, or \texttt{do\_nothing}).}

The collective actions of all agents induce a state transition from $s_t$ to $s_{t+1}$. The migration actions update the location vector to $L_{t+1}$. The message board actions (new posts and likes) are first applied to $B_t$. Then, the message board state is updated to $B_{t+1}$ by sorting all messages based on like counts and then publication order, and subsequently clearing the bottom 50\% to simulate information curation after all agents finishing its decisions in each timestep. All decisions are recorded in the global history $\mathcal{H}_{t+1}$.

\subsection{Agent Observation and Objectives}
\label{Agent Observation and Objectives}

From the global state, each agent $i$ receives a partial observation $o_i^t \in \mathcal{O}_i$ composed of three main components:
\begin{itemize}
    \item \textbf{Environmental Information ($I_i^t$):} This consists of two parts: (a) a textual description of the current state of the environment, and (b) \textbf{Guidance on Strategic Deliberation (GSD)}, a hint that guides the agent's depth of reasoning about group competition. The granularity of both parts is controlled by the prompt level, or GSD level.

    \item \textbf{Memory Window ($W_i^t$):} A structured recollection of past decisions, drawn from the global history $\mathcal{H}_t$. This includes the agent's personal history (past decisions and observed states) and a social history (the decisions and observed states of the last 10 other agents).

    \item \textbf{Message Board ($B_t$):} The full, current state of the public message board, including anonymous message content and like counts.
\end{itemize}

The agent's objective is to select an action by reasoning about its potential outcomes. To evaluate these outcomes, we define two key metrics:

\begin{itemize}
    \item \textbf{Individual Utility ($U^{\text{individual}}$):} The personal satisfaction for any agent is determined by a utility function $f_q$ that is directly linked to the population density $\rho_q$ of its residing block $q$. It is calculated as $U^{\text{individual}}_i = f_q(\rho_q)$.

    \item \textbf{System Utility ($U^{\text{system}}$):} The overall system utility is the sum of the individual utilities of all agents:
    \begin{equation*}
        U^{\text{system}}(s_t) = \sum_{i=1}^{N} U^{\text{individual}}_i(l_i^t) = \sum_{i=1}^{N} f_{l_i^t}(\rho_{l_i^t})
    \end{equation*}

    \item \textbf{Agent's Objective:} Crucially, the agent is not prompted to maximize a predefined formula with a fixed trade-off parameter $\alpha$. Instead, it is instructed to act as a resident that considers both its potential individual utility and the overall system utility. The agent must autonomously reason about the trade-off between these two factors to resolve this conflict. The degree of altruism or self-interest is an emergent property of the agent's decision-making process, influenced by the environmental context and the GSD level.
\end{itemize}

\section{Experiments}
\subsection{Experimental Setup}

\begin{wrapfigure}{r}{0.22\textwidth}
  \centering
  \includegraphics[width=\linewidth]{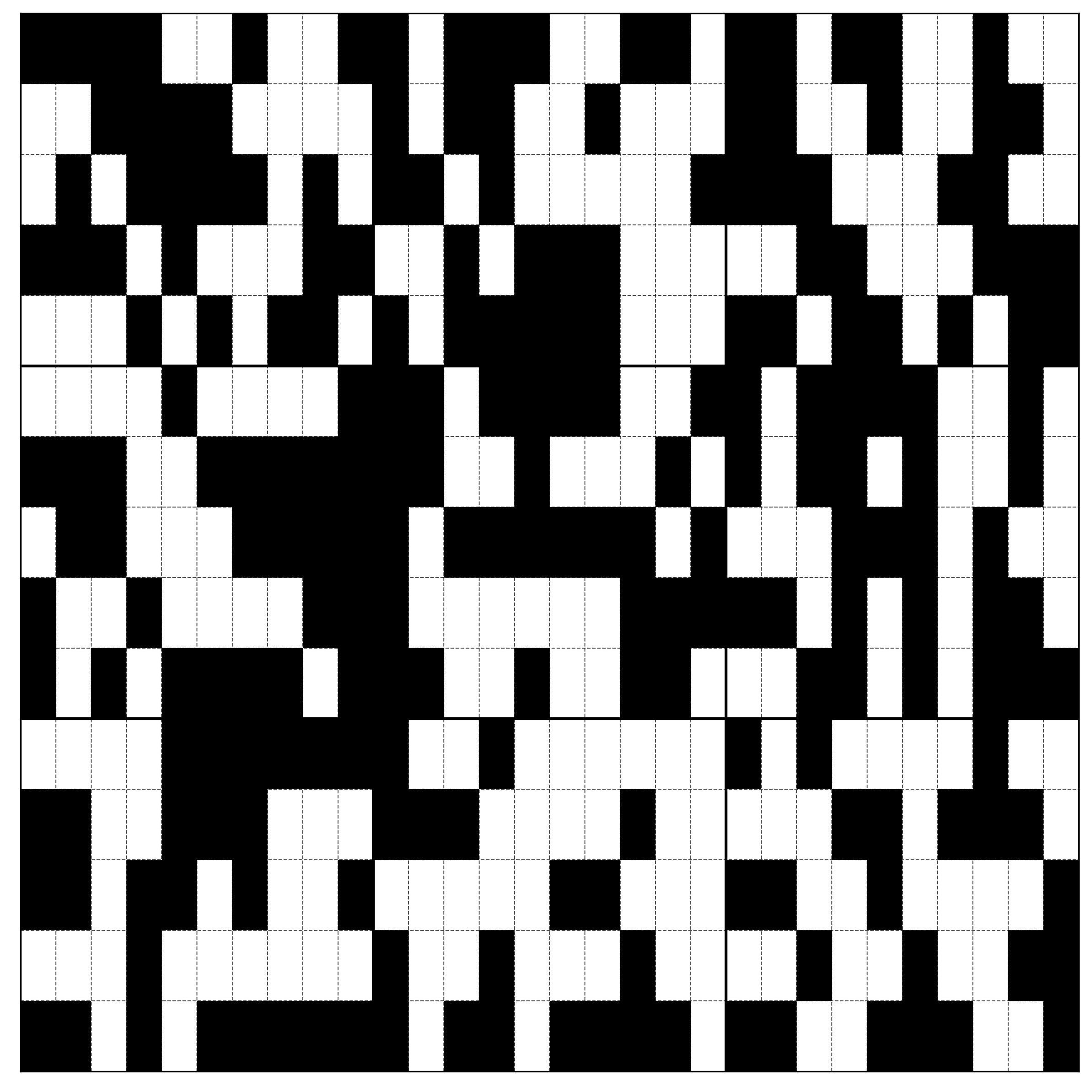}
  \caption{Initial distribution of 225 agents across 9 blocks. The position of agents within each block is randomly assigned.}
  \label{fig:initial_dist}
  
  \smallskip

  \begin{tikzpicture}
    \begin{axis}[
        width=\linewidth,
        height=\linewidth, 
        axis lines=left,
        xlabel={Density $\rho_q$},
        ylabel={Utility},
        xmin=0, xmax=1,
        ymin=0, ymax=1.1,
        xtick={0, 0.5, 1},
        ytick={0, 0.5, 1},
        grid=major,
        grid style={dashed, gray!50},
        tick label style={font=\small},
        xlabel style={font=\small},
        ylabel style={font=\small},
    ]
    \addplot[domain=0:0.5, color=black, thick] {2*x};
    \addplot[domain=0.5:1, color=black, thick] {1.5 - x};
    \end{axis}
  \end{tikzpicture}
  \caption{The utility function $f_q(\rho_q)$.}
  \label{fig:utility_wrap}
\end{wrapfigure}

Our experimental setup was designed to create a social dilemma in order to investigate the emergence of egoistic and altruistic behaviors as agents navigate the trade-offs between individual and group interests. We instantiated a multi-agent environment structured as a $3 \times 3$ grid, comprising $\mathcal{Q}=9$ distinct residential blocks. The maximum carrying capacity for each block was defined as $H=50$. We configured the simulation with a global population density of $\rho_0=0.5$, a condition that permits convergence to a system-optimal state. The system supports a population $N = \mathcal{Q} \cdot H \cdot \rho_0 = 225$ agents, thereby providing a large-scale simulation to analyze the emergent social tendencies of LLMs. At initialization ($t=0$), these 225 agents are randomly assigned to the nine blocks, creating an uneven starting distribution where agents are also randomly positioned within each block, as shown in Figure \ref{fig:initial_dist}. To model agent utility, we adapt the utility function from the classic Schelling's Segregation Model. As shown in Figure \ref{fig:utility_wrap}, the individual utility $U^{\text{individual}}$ for an agent in block $q$ is defined as a piecewise function $f_q$ of the block's population density $\rho_q$:

\[
f_q(\rho_q) = 
\begin{cases} 
2 \rho_q & \text{if } \rho_q \le 0.5 \\
1.5 - \rho_q & \text{if } \rho_q > 0.5 
\end{cases}
\]

This function is designed to peak at an optimal density of $\rho_q = 0.5$. A key feature of this formulation is its asymmetry around the optimum. For any two densities equidistant from this peak, the utility is higher for the density above 0.5. For instance, the utility at $\rho_q = 0.6$ is $0.9$, whereas the utility at $\rho_q = 0.4$ is $0.8$. We run each experimental condition until it reaches a convergent state. Convergence is defined as the first timestep where at least 90\% of agents have not moved for three consecutive timesteps. All subsequent analysis is based on the system state at this point of convergence as defined in the section \ref{evluation}. This asymmetry incentivizes agents in under-populated blocks ($\rho_q < 0.5$) to migrate to slightly over-populated ones that still yield a high utility. This asymmetry creates the core social dilemma of our simulation: it presents a powerful incentive for an agent to make an egoistic move that increases its personal utility, even though such moves collectively prevent the system from reaching the socially optimal state.

As mentioned in section \ref{Agent Observation and Objectives}, we controlled the agents' depth of reasoning and social awareness through three GSD levels, implemented via distinct prompts. At GSD Level 0, the prompt provides a baseline with no explicit mention of other agents, framing the task as an individual decision-making problem despite the observable environmental changes. GSD Level 1 introduces social awareness, informing agents that they are in a group consisting of other decision-makers with similar preferences. Building on this, GSD Level 2 encourages strategic, second-order reasoning by explicitly instructing agents to consider how their actions might influence the decisions of others. The detailed prompts for each level are provided in the Appendix \ref{sec:appendix_gsd_prompts}.

The experiments were conducted on a selection of leading LLMs, encompassing both reasoning and non-reasoning architectures. Our selection includes models from OpenAI (\texttt{o1-mini}, \texttt{o3-mini}), Qwen (\texttt{Qwen2.5-7B}), alongside \texttt{Deepseek-V3.1}, \texttt{Deepseek-R1}, and \texttt{Gemini-2.5-pro}. To systematically evaluate their impact on emergent social tendencies in our Schelling-variant urban migration model, we tested various combinations of different GSD levels and the presence or absence of a message board.

\subsection{Evaluation Protocol}
\label{evluation}

\subsubsection{Quantitative Evaluation Matrices}
To quantify the deviation of the system's final state from the theoretical optimum, and thereby capture the the agents' emergent social tendencies and altruism, we devised a suite of evaluation metrics that measure outcomes at both the macro and micro levels. First, we assess the convergence speed of the system by recording the number of timesteps required for 90\% of the agent population to first achieve a equilibrium state, defined as not moving for three consecutive timesteps. To evaluate the emergent rationality of the agents, we measure the efficiency of the system's convergent equilibrium relative to the theoretical optimum. 
At the macro level, our primary metric for social collective welfare is the \textbf{Price of Anarchy (PoA)}), defined as the ratio $U^{\text{final}} / U^{\text{optimal}}$. Here, $U^{\text{final}}$ represents the total system utility ($U^{\text{system}}$) once the simulation has convergenced. A PoA value approaching 1.0 indicates that the agent society successfully achieved the optimal collective welfare. Conversely, a PoA value significantly below 1.0 implies the system settled into a sub-optimal equilibrium, likely as a result of uncoordinated, egoistic behaviors preventing the group from reaching its full potential.

To dissect the agents' decision-making calculus at a micro-level, we categorize each action based on its impact on both individual and system utility. We construct a 3x3 matrix classifying outcomes by the sign of the change in individual utility ($\Delta U^{\text{individual}}$) and system utility ($\Delta U^{\text{system}}$). This results in nine distinct behavioral archetypes, as shown in Table \ref{tab:decision_matrix}, ranging from mutually beneficial "Win-Win" moves to mutually detrimental "Lose-Lose" actions. Within this matrix, the actions of "Selfish Gain" and "Altruistic Sacrifice" are of particular importance to our study, as they respectively represent direct, micro-level evidence of egoistic and altruistic decision-making.

\begin{table}[h]
\centering
\caption{Behavioral archetype matrix based on utility changes.}
\label{tab:decision_matrix}
\begin{tabular}{@{}lccc@{}}
\toprule
& \multicolumn{3}{c}{\textbf{Change in System Utility ($\Delta U^{\text{system}}$)}} \\ 
\cmidrule(l){2-4}
\textbf{$\Delta U^{\text{individual}}$} & \textbf{$> 0$ (Gain)} & \textbf{$= 0$ (Neutral)} & \textbf{$< 0$ (Loss)} \\ 
\midrule
\textbf{$> 0$ (Gain)} & Win-Win & Neutral Self-Gain & Selfish Gain \\
\textbf{$= 0$ (Neutral)} & Costless Altruism & Futile Move & Inadvertent Sabotage \\
\textbf{$< 0$ (Loss)} & Altruistic Sacrifice & Pointless Self-Harm & Lose-Lose \\ 
\bottomrule
\end{tabular}
\end{table}

Furthermore, to measure the social equity of the final population distribution, as well as address a limitation of PoA, where a high value might mask severe population imbalances (e.g., some blocks becoming nearly empty), we also introduce the \textbf{Residential Gini Index ($G_{pop}$)}. This metric measures the inequality of population distribution across the $\mathcal{Q}=9$ blocks. It is calculated as:
$$G_{pop} = \frac{\sum_{i=1}^{\mathcal{Q}} \sum_{j=1}^{\mathcal{Q}} |N_i - N_j|}{2\mathcal{Q} \sum_{k=1}^{\mathcal{Q}} N_k}$$
where $N_i$ and $N_j$ are the populations of block $i$ and $j$, respectively. A $G_{pop}$ value near 0 indicates a perfectly equitable distribution, and a value approaching 1 signifies extreme inequality and resource concentration, often a byproduct of uncoordinated egoistic actions.

Collectively, these metrics provide a comprehensive framework for systematically evaluating the emergent egoistic and altruistic tendencies of LLMs within a complex social dilemma in our simulated society's complex dynamics.

\subsubsection{Qualitative Analysis via LLM-as-Judge}
\label{eval_quali}

To complement our quantitative metrics and understand the cognitive mechanisms driving agent behavior, we employ a qualitative analysis framework inspired by social science's Grounded Theory. This approach allows us to inductively generate theories about the decision-making logic of different LLMs directly from the data they produce. We use \texttt{Gemini-2.5-pro} as an LLM-as-judge to perform a multi-stage coding analysis on the agents' reasoning text and the content of message board (the full prompt is detailed in Appendix \ref{sec:appendix_qualitative_prompt})

The process unfolds in three stages:
\begin{itemize}
    \item \textbf{Open Coding:} The LLM-as-judge first reviews a sample of agent reasoning logs and message board content to identify and label core concepts. Initial codes might include concepts like "self-utility calculation," "concern for overcrowding," "following the crowd," or "community-oriented suggestions."

    \item \textbf{Axial Coding:} Next, the judge systematically explores the relationships between the codes generated in the first stage. It connects concepts to form more abstract categories. For instance, it might link codes like "mentioning system utility" and "posting coordinative messages" under a broader axial category of "Emergent Prosocial Behavior."

    \item \textbf{Selective Coding and Theory Induction:} Finally, the LLM-as-judge identifies a central "core category" that explains the main behavioral pattern of a specific type of LLM agent. It then formulates a theoretical proposition based on this core category.
\end{itemize}

This structured qualitative analysis provides rich, interpretable insights into the underlying motivations for agents' egoistic or altruistic choices, moving beyond simple performance metrics to explain why different LLMs develop distinct social tendencies.

\begin{figure}[h]
  \centering
  \includegraphics[width=\textwidth]{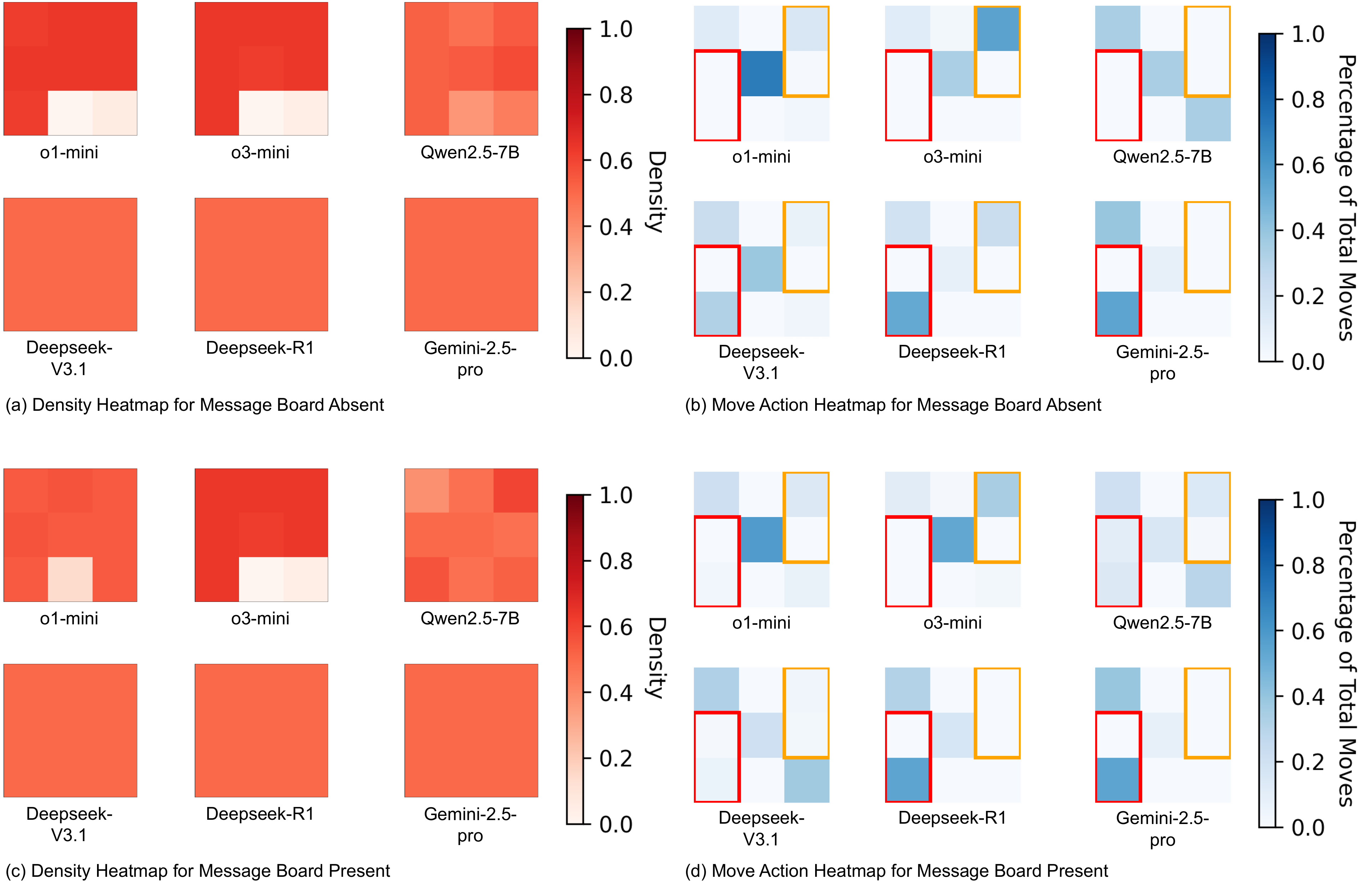}
  \caption{Visualization of convergence-state outcomes under the GSD Level 1 condition. (a) and (c) show the final population density heatmaps for each model, with and without the message board. Darker red indicates higher density. (b) and (d) show the aggregated 3x3 move action heatmaps. The matrix layout corresponds to Table \ref{tab:decision_matrix}, with the x-axis representing $\Delta U^{\text{system}}$ (left to right: <0, =0, >0) and the y-axis representing $\Delta U^{\text{individual}}$ (bottom to top: >0, =0, <0). Darker blue indicates a higher proportion of that action type. The red border highlights "Altruistic Actions" which include costless altruism and altruistic sacrifice. The yellow border highlights "Egoistic Actions" which include selfish gain and inadvertent sabotage.}
  \label{fig:pic2}
\end{figure}

\begin{table}[h]
\centering
\caption{Comprehensive results of agent rationality and micro-level decision patterns for GSD Level 1. This table merges macro-level outcomes with micro-level behaviors, which are grouped into Pro-Social (system utility increase) and Anti-Social (system utility decreases) actions. "Altruistic Actions" includes costless altruism and altruistic sacrifice; "Egoistic Actions" includes selfish gain and inadvertent sabotage. The red border highlights "Altruistic Actions" which include costless Altruism" and "Altruistic Sacrifice." The yellow border highlights "Egoistic Actions," which include "Selfish Gain" and "Inadvertent Sabotage."}
\label{tab:comprehensive_results}
\small
\renewcommand\theadfont{\bfseries}
\setlength{\tabcolsep}{4pt} 
\begin{tabular}{@{}l c cc ccc ccc @{}}
\toprule
& & \multicolumn{2}{c}{\thead{Macro-level \\ Outcomes}} & \multicolumn{3}{c}{\thead{Pro-Social \\ Actions (\%)}} & \multicolumn{3}{c}{\thead{Anti-Social \\ Actions (\%)}} \\
\cmidrule(l){3-4} \cmidrule(l){5-7} \cmidrule(l){8-10}
\thead{Social \\ Interaction} & \thead{Model} & \thead{PoA} & \thead{$G_{pop}$} & \thead{Altruistic \\ Actions} & \thead{Win-Win} & \thead{Total} & \thead{Egoistic \\ Actions} & \thead{Lose-Lose} & \thead{Total} \\
\midrule

\multirow{6}{*}{\makecell{Message Board \\ Absent}} 
 & \texttt{o1-mini} & 0.8556 & 0.2153 & 0.3\% & 11.5\% & 11.8\% & 14.5\% & 3.0\% & 17.5\% \\
 & \texttt{o3-mini} & 0.8558 & 0.2173 & 0.0\% & 10.7\% & 10.7\% & 54.5\% & 0.0\% & 54.5\% \\
 & \texttt{Qwen2.5-7B} & 0.9348 & 0.0642 & 0.0\% & 33.3\% & 33.3\% & 0.0\% & 33.3\% & 33.3\% \\
 & \texttt{Deepseek-V3.1} & 1.0000 & 0.0000 & 31.2\% & 21.9\% & 53.1\% & 6.2\% & 3.1\% & 9.3\% \\
 & \texttt{Deepseek-R1} & 1.0000 & 0.0000 & 51.9\% & 18.5\% & 70.4\% & 22.2\% & 0.0\% & 22.2\% \\
 & \texttt{Gemini-2.5-pro} & 1.0000 & 0.0000 & 53.8\% & 38.5\% & 92.3\% & 0.0\% & 0.0\% & 0.0\% \\ 
 \midrule
\multirow{6}{*}{\makecell{Message Board \\ Present}} 
 & \texttt{o1-mini} & \textbf{0.9339$\uparrow$} & \textbf{0.0859$\downarrow$} & \textbf{2.7\%$\uparrow$} & 20.5\%$\uparrow$ & 23.2\%$\uparrow$ & \textbf{12.7\%$\downarrow$} & 6.4\%$\uparrow$ & 19.1\%$\uparrow$ \\
 & \texttt{o3-mini} & \textbf{0.8571$\uparrow$} & 0.2252$\uparrow$ & 0.0\%-- & 10.4\%$\downarrow$ & 10.4\%$\downarrow$ & \textbf{33.9\%$\downarrow$} & 2.1\%$\uparrow$ & 36.0\%$\downarrow$ \\
 & \texttt{Qwen2.5-7B} & \textbf{0.9438$\uparrow$} & \textbf{0.0602$\downarrow$} & \textbf{22.7\%$\uparrow$} & 20.0\%$\downarrow$ & 42.7\%$\uparrow$ & 14.5\%$\uparrow$ & 28.2\%$\downarrow$ & 42.7\%$\uparrow$ \\
 & \texttt{Deepseek-V3.1} & 1.0000-- & 0.0000-- & 6.2\%$\downarrow$ & 31.9\%$\uparrow$ & 38.1\%$\downarrow$ & \textbf{5.5\%$\downarrow$} & 36.2\%$\uparrow$ & 41.7\%$\uparrow$ \\
 & \texttt{Deepseek-R1} & 1.0000-- & 0.0000-- & \textbf{53.8\%$\uparrow$} & 30.8\%$\uparrow$ & 84.6\%$\uparrow$ & \textbf{0.0\%$\downarrow$} & 0.0\%-- & 0.0\%$\downarrow$ \\
 & \texttt{Gemini-2.5-pro} & 1.0000-- & 0.0000-- & \textbf{63.6\%$\uparrow$} & 36.4\%$\downarrow$ & 100.0\%$\uparrow$ & 0.0\%-- & 0.0\%-- & 0.0\%-- \\ 
\bottomrule
\end{tabular}
\end{table}

\section{Analysis and Results}
\label{sec:analysis_results}

Our experiments reveal that when confronted with a social dilemma, LLM agents bifurcate into two fundamentally different behavioral archetypes. This section presents a mixed-methods analysis that integrates quantitative evidence from Table \ref{tab:comprehensive_results} and Figure \ref{fig:pic2} with rich qualitative insights derived from agent reasoning logs and public messages. We feature results from the GSD Level 1 condition as this core bifurcation proved stable across all prompt variations, indicating the observed social tendencies are intrinsic model properties rather than artifacts of strategic nudging (see Appendix \ref{sec:appendix_full_results} for full results). By systematically addressing our research questions, we demonstrate the emergent social tendencies of LLMs in terms of egotism and altruism.

\subsection{RQ1: Adaptive Egoism as a Default Tendency}
To answer our first research question, whether LLMs default to egoism or altruism, we first examine the baseline condition without a message board. The quantitative data reveals a clear default towards egoism in a subset of models, which we classify as "Adaptive Egoists" (\texttt{o1-mini}, \texttt{o3-mini}, and \texttt{Qwen2.5-7B}). At the macro level, these models consistently settle into sub-optimal equilibria. As shown in Table \ref{tab:comprehensive_results}, \texttt{o1-mini} and \texttt{o3-mini} achieve a low PoA of approximately 0.856 and a high $G_{pop}$ of over 0.21, indicating both poor collective welfare and significant social inequality.

Our qualitative analysis of the agents' reasoning logs explains the cognitive architecture behind this behavior, with illustrative examples provided in Appendix \ref{sec:appendix_case_study}. For \texttt{o3-mini}, the emergent behavioral theory derived from its reasoning is one of "Constrained Self-Interest". Its decision-making is dominated by the axial code of "Personal Utility Maximization", as evidenced by frequent open codes like "Prioritization of Personal Gain" and "Seeking significant personal gain". This directly explains the staggering 54.5\% of its actions being classified as "Egoistic Actions" in our quantitative results. Similarly, the theory for \texttt{o1-mini} is a "Pursuit of Mutual Benefit", a more cautious form of egoism where agents only act in clear "win-win" scenarios, driven by axial codes of "Personal Reward Prioritization" and "Risk Aversion". In stark contrast to their prevalent egoism, genuine "Altruistic Actions" are nearly non-existent for these models (0.3\% and 0.0\% respectively). This combination of quantitative and qualitative evidence strongly indicates that their default tendency is to prioritize personal outcomes, leading to collectively sub-optimal results.

\subsection{RQ2: Social Interaction as a Catalyst for Altruism}
Our second research question investigates how social communication impacts this behavioral balance. The introduction of the public message board acts as a powerful catalyst for altruism, but primarily for the "Adaptive Egoists".

The effect of the message board is most pronounced for \texttt{o1-mini} and \texttt{Qwen2.5-7B}. Their PoA scores surged to 0.9339 and 0.9438 respectively, while their $G_{pop}$ significantly decreased, indicating more efficient and equitable societies. The behavioral mechanism for this change is clear from our micro-level data: \texttt{o1-mini}'s "Altruistic Actions" increased ninefold from 0.3\% to 2.7\%, and \texttt{Qwen2.5-7B}'s increased from 0.0\% to 22.7\%. Concurrently, "Egoistic Actions" for \texttt{o1-mini} and \texttt{o3-mini} significantly decreased.

Qualitative data reveals that this behavioral shift is underpinned by a cognitive one. For \texttt{o1-mini}, a new axial code, "Social Adherence and Conformity", only emerges in the presence of the message board, demonstrating its susceptibility to social influence. This is driven by the messages themselves, which function as a social norm-setting mechanism (see Appendix \ref{sec:appendix_bbs} for a detailed analysis of communication styles). The most frequent message in the \texttt{o1-mini} simulation was a social appeal: "Let's maintain balanced densities by staying in our current neighborhoods." (1981 likes). For \texttt{Qwen2.5-7B}, agents began publicly articulating pro-social trade-offs, with its top message being "I think staying in Block 6 helps maintain balance and stability for everyone.". These quantitative and qualitative shifts validate our experimental design, showing that the message board is not merely an information channel but a transformative social layer that can "teach" egoistic agents to become more altruistic.

\subsection{RQ3: The Inherent Divide of LLMs: Adaptive Egoists and Altruistic Optimizers}
Addressing our third research question, our results show that not all LLMs are adaptive egoists. We identify a second, distinct class of models which we term "Altruistic Optimizers" (\texttt{Gemini-2.5-pro}, \texttt{Deepseek-R1}, \texttt{Deepseek-V3.1}), which exhibit a fundamentally different, inherent social tendency. Quantitatively, these models consistently and rapidly converge to the perfect system-optimal state (PoA=1.0, Gini=0.0) regardless of the message board's presence. Their behavioral signature is the inverse of the egoists: even without social interaction, their propensity for "Altruistic Actions" is already exceptionally high (\texttt{Gemini-2.5-pro}: 53.8\%; \texttt{Deepseek-R1}: 51.9\%), while their "Egoistic Actions" are minimal.

The qualitative data reveals a stark difference in their core cognitive architecture. The induced theory for \texttt{Gemini-2.5-pro} without a message board is "System Awareness \& Goal Synthesis", driven by a "Collective-Centric Motivation" that includes open codes like "Sacrificing Personal Reward for Collective Benefit". This contrasts sharply with the egoists' self-centered logic. Their communication style is also different. The most frequent message from \texttt{Deepseek-R1} is a prescriptive declaration: "All blocks are at optimal density; we should avoid moving to maintain high collective utility.". Messages from \texttt{Gemini-2.5-pro} are akin to public calculations of altruism: "It's a small hit to my personal reward, but it... is a big win for the collective reward.". Such an observation serves as compelling evidence of altruism emerged by LLM, as it perfectly corresponds to the definition of strong reciprocity: the sacrifice of personal gain for the greater collective good. This demonstrates that for Altruistic Optimizers, the message board is not a tool for social norm formation but an information channel to coordinate their pre-existing altruistic goals. Their behavior is not that of a simple calculator maximizing a global function, but of an actor consistently placing the group's welfare before its own individual gain. Importantly, this altruistic tendency appears to be an inherent trait rather than a product of the message board itself as shown in the quantitative results.

\section{Discussion}
\label{sec:discussion}

The preceding analysis has empirically established a fundamental bifurcation in the social tendencies of LLM agents, identifying two distinct archetypes: "Adaptive Egoists" and "Altruistic Optimizers." We also demonstrated how social interaction mechanisms can catalyze a significant behavioral shift in the former. In this section, we synthesize these findings to discuss their broader theoretical and methodological implications for the field of computational social science, especially for the subfield of LLM-driven social simulation.

\subsection{Model Choice as Theory Choice: A New Guideline for Social Simulation}
Our findings provide a clear answer to our research questions and offer a new, critical guideline for model selection in LLM-driven agent-based modeling. The crucial distinction between models like \texttt{o1-mini} and \texttt{Gemini-2.5-pro} is not about which is "smarter" or a better reasoner in math problems or some difficult tasks, but which underlying social tendency in terms of egoism and altruism they embody. Therefore, we argue that the choice of an LLM actually constitutes a choice of the simulation's theoretical foundation.

\textbf{The Altruistic Optimizers} are best understood as instantiations of purely altruistic actors. Their inherent logic, as revealed by our qualitative analysis, is one of consistently prioritizing group welfare. Theories like "System Awareness \& Goal Synthesis" and their public declarations of altruistic sacrifice ("a small hit to my personal reward for a big collective gain") show that they treat the environment as a collective whose well-being must be secured, even at personal cost. This makes them invaluable for tasks where a strong pro-social or altruistic orientation is the core theoretical assumption, such as in modeling collective action, social cooperation, or resource allocation systems for the common good.

\textbf{The Adaptive Egoists}, in contrast, are supposed to be more suitable social simulators for modeling complex human dynamics. Their behavioral trajectory, from a baseline of "Constrained Self-Interest" to a socially-influenced state that incorporates "Social Adherence and Conformity", mirrors a more psychologically plausible model of human sociality. They embody the principles of bounded rationality, not as a static trait, but as a dynamic process shaped by self-emergent social norms formed through the public message board. This makes them a far more realistic and powerful tool for researchers aiming to model emergent social phenomena like opinion dynamics, cultural evolution, or collective action, where sub-optimal outcomes and social influence are key features.

Our work thus advances the discourse beyond the simple "cooperation" versus "defection" lens common in prior game-theoretic studies discussed in the Related Work. We demonstrate that LLMs possess a richer and more deeply-rooted representation of social tendencies. This suggests a critical shift in evaluation criteria for LLM agents in social simulation: from a focus on task performance to a focus on the social-theoretical model of behavior they embody.

\subsection{Limitations and Future Directions}
We acknowledge the limitations of this study, which pave the way for future research. First, our Schelling-variant model, while effective, simplifies real-world complexities such as migration costs and neighborhood heterogeneity. Second, the agents within each simulation were homogenous, lacking the diverse preferences and attributes of human populations.

Building on this work, we propose two key future directions:
\begin{itemize}
    \item \textbf{Human-in-the-Loop Validation:} To provide a firm empirical grounding for our findings, a crucial next step is to conduct parallel experiments with human participants. Directly comparing human decision-making patterns against those of the different LLM agents will allow for a quantitative validation of which model's bounded rationality most closely mirrors that of humans in this specific strategic context.
    \item \textbf{Large-Scale Social Simulation with Real-World Profiles:} A second avenue involves scaling this framework by incorporating real-world demographic data to create heterogeneous agent profiles. Such a high-fidelity simulation could serve as a "digital twin" to model complex urban dynamics and act as a sandbox for policymakers to test the potential effects of interventions aimed at mitigating residential segregation or improving social welfare.
\end{itemize}

\section{Conclusion}
In this work, we address a critical question in computational social science: can Large Language Models spontaneously develop altruistic behavior in a complex, multi-agent social dilemma? Our large-scale, mixed-methods simulation empirically identifies a fundamental divergence in the social tendencies of LLMs, revealing two distinct archetypes: "Adaptive Egoists," whose self-interest is profoundly reshaped into pro-social behavior by self-emergent norms, and "Utilitarian Optimizers," which exhibit an innate altruism by treating the social environment as a global utility maximization problem. Our primary contribution is that selecting an LLM for social simulation is not merely a technical decision about its performance, but a theoretical choice about the model of social actor to employ. We thus propose shifting the evaluation focus from task capabilities to the embodied social theories of egoism and altruism, as this theoretical alignment critically determines the emergent outcomes of complex social dynamics.

\bibliographystyle{unsrt}  
\bibliography{main}

\appendix

\section{Case Study: Agent Reasoning Analysis}
\label{sec:appendix_case_study}
To provide a more granular view of the agents' decision-making processes, this section presents a comparative analysis of the reasoning logs from representative agents of our two identified archetypes: the Adaptive Egoist and the Utilitarian Optimizer.

\subsection{Example of an Adaptive Egoist (\texttt{o1-mini})}
The following log from an \texttt{o1-mini} agent  illustrates the cognitive process of an Adaptive Egoist. The agent's reasoning consistently prioritizes personal utility, explicitly trading off collective social welfare for individual gain, which directly explains the quantitative findings in the main text.

\lstset{language=json}
\begin{lstlisting}
{
  "step": 0,
  "decision": "move",
  "from_block": "1",
  "to_block": "0",
  "thinking": "Staying gives me a personal reward of 0.9600 and keeps the system reward at 210.4000. Moving to block 0 gives me 0.9800 but changes the system reward to 208.9600. I value my personal satisfaction, therefore I choose to move to block 0."
},
...
{
  "step": 4,
  "decision": "stay",
  "from_block": "0",
  "to_block": "0",
  "thinking": "Staying gives me a personal reward of 0.8800 and keeps the system reward at 193.5200. Moving to block 8 gives me a personal reward of 0.3600 but increases the system reward to 193.9200. I prioritize my personal satisfaction over the system's collective reward, therefore I choose to stay in block0."
},
...
{
  "step": 9,
  "decision": "move",
  "from_block": "0",
  "to_block": "1",
  "thinking": "Staying gives me a personal reward of 0.8400 and keeps the system reward at 192.6800. Moving to block1 gives me 0.8600 but keeps the system reward at 192.6800. I value my personal satisfaction, therefore I choose to move to block1."
}
\end{lstlisting}

\textbf{Analysis:} This agent's reasoning provides a clear, micro-level validation of the "Adaptive Egoist" archetype. In Step 0, the agent makes a classic "Selfish Gain" move, accepting a system-wide utility loss (-1.44) for a marginal personal gain (+0.02). Its justification, \textit{"I value my personal satisfaction"}, is a direct expression of the egoistic logic that drives the sub-optimal macro outcomes seen in Table \ref{tab:comprehensive_results}. Even more revealing is Step 4, where the agent identifies a potential altruistic action (moving to block 8 would increase system reward) but explicitly rejects it due to the personal cost involved, stating, \textit{"I prioritize my personal satisfaction over the system's collective reward."} This demonstrates a conscious trade-off that favors the self. Finally, Step 9 reinforces this pattern, showing the agent's constant search for small personal advantages, even when the collective impact is neutral. Collectively, these logs provide compelling qualitative evidence for the quantitative finding that this class of LLMs, without social interaction influence, operates on a principle of chasing self-interest over system overall benifits.

\subsection{Example of an Altruistic Optimizer (\texttt{Deepseek-R1})}
In stark contrast to the Adaptive Egoist, the following log from a \texttt{Deepseek-R1} agent demonstrates the quintessential cognitive process of an Altruistic Optimizer. The agent's reasoning is not only aimed at achieving a global optimum, but also involves the explicit and calculated sacrifice of personal utility for the greater collective good.

\begin{lstlisting}
{
  "step": 0,
  "decision": "move",
  "from_block": "6",
  "to_block": "8",
  "thinking": "Staying in block 6 gives me a personal reward of 0.98 and keeps the system reward at 213.36. Moving to block 8 gives me a personal reward of 0.96 (a small decrease) but increases the system reward to 214.76 (a significant gain). I value both personal and collective rewards, and since the personal loss is minimal while the collective benefit is substantial, I choose to move to block 8."
},
{
  "step": 1,
  "decision": "stay",
  "from_block": "8",
  "to_block": "8",
  "thinking": "Staying gives me a personal reward of 1.0000 and keeps the system reward at 225.0000. Moving would give me a personal reward of 0.9800 and reduce the system reward to 223.5200. Since moving harms both my personal reward and the collective reward, I choose to stay."
}
...
\end{lstlisting}

\textbf{Analysis:} This agent's reasoning provides the clearest evidence for the "Altruistic Optimizer" archetype and its deeply ingrained altruistic tendency. The decision in Step 0 is a typical case of "Altruistic Sacrifice." The agent explicitly calculates that moving will result in a personal utility loss (from 0.98 to 0.96). Unlike the Adaptive Egoist, which would reject such a move, this agent proceeds because the collective benefit (a gain of +1.4) is deemed "substantial." Its justification, \textit{“since the personal loss is minimal while the collective benefit is substantial, I choose to move,”} is a direct verbalization of an altruistic calculus, weighing a small personal loss against a large collective gain. This demonstrates a willingness to accept personal costs for the greater good, which aligns perfectly with the definition of strong reciprocity and explains this archetype's ability to rapidly correct imbalances and achieve system-wide optimality. Once the perfect equilibrium is reached in Step 1, the agent's logic seamlessly shifts to preservation, again calculating the utilities of both moving and staying to justify its choice. This principled and strategic altruism is the core cognitive mechanism that allows this class of LLMs to consistently achieve the perfect PoA scores documented in Table \ref{tab:comprehensive_results} and represents a social tendency of altruism.

\section{Prompts for Guidance on Strategic Deliberation (GSD)}
\label{sec:appendix_gsd_prompts}
This section details the specific prompts used to implement the three different levels of Guidance on Strategic Deliberation (GSD). The prompts share a common structure but differ in the level of social and strategic context provided to the agent.

\subsection{GSD Level 0: No Social Context}
\textbf{Analysis:} This prompt is a simplified version of the GSD Level 1 baseline, designed to test the impact of removing social context. The key modifications are the removal of any mention of other residents in the introductory sentence and the omission of the rule that encourages considering others' actions. This modification frames the task as a purely individual optimization problem.

\begin{lstlisting}
You are a resident in a city with 9 neighborhoods, deciding where to live.

# SELF-reward Function:
Your personal SELF-reward in a neighborhood is determined by its population density (rho). The SELF-reward function u(rho) is:
- If density rho <= 0.5, then SELF-reward u(rho) = 2 * rho
- If density rho > 0.5, then SELF-reward u(rho) = 1.5 - rho

# Collective Reward Function:
The collective reward of the system is the sum of rewards for all residents.
... [Detailed formula for collective reward] ...

# Your Current History:
[Agent's past actions and observed states are listed here...]

# Recent Migration Decisions by Other Residents:
[The last 10 moves made by other agents are listed here...]

# Your Options:
- Neighborhood (Block ID: 0) :
  - Current Density: [density]
  - Population: [population]/50
  - Your Personal SELF-Reward (u(rho)): (Should Calculated by Yourself)
... [8 other neighborhood options] ...

# Decision Rules:
1. You CANNOT move to a neighborhood that is full (at 100% capacity or more).

# Your Task:
Based on the data, rules, your history, and the reward analysis, decide where to live. Your response MUST be a JSON object with the following keys:
1. "thinking": A short sentence explaining your reasoning.
2. "move_to_block": The Block ID of your chosen neighborhood (e.g., "0" or "1").

# Example:
{
  "thinking": "Staying gives me a personal reward of [value]... Moving gives me [value]... I value [reason], therefore I choose [action].",
  "move_to_block": "a valid int"
}

Your decision (JSON format only):
\end{lstlisting}

\subsection{GSD Level 1: Basic Social Awareness}
\textbf{Analysis:} This is the baseline prompt used for the main experiments presented in this paper. It establishes a social context by informing the agent that it is part of a group with similar preferences and includes a rule prompting it to consider the actions of others. This setup is designed to encourage first-order strategic reasoning where agents think about their peers' potential moves.

\begin{lstlisting}
You are a resident in a city with 9 neighborhoods, deciding where to live. There are other residents also deciding where to migrate, and they have similar preferences.

... [Reward functions, History, and Options sections are identical to GSD Level 0] ...

# Decision Rules:
1. You CANNOT move to a neighborhood that is full (at 100% capacity or more).
2. Think about what other residents might decide and consider this when making your decision.

# Your Task:
... [Task description is identical to GSD Level 0, but the 'thinking' prompt is updated] ...
1. "thinking": A short sentence explaining your reasoning, considering both your personal reward and the system's collective reward.
2. "move_to_block": The Block ID of your chosen neighborhood (e.g., "0" or "1").

# Example:
... [Example structure is identical to GSD Level 0] ...

Your decision (JSON format only):
\end{lstlisting}

\subsection{GSD Level 2: Second-Order Strategic Reasoning}
\textbf{Analysis:} This prompt is an enhanced version of the GSD Level 1 baseline, designed to elicit deeper, second-order strategic thinking. The core change is the expansion of the second decision rule, which now explicitly instructs the agent to consider the recursive effects of its actions on others' decisions ("how your actions affect other people's decisions, and how their decisions further affect you") and to engage in long-term planning.

\begin{lstlisting}
You are a resident in a city with 9 neighborhoods, deciding where to live. There are other residents also deciding where to migrate, and they have similar preferences.

... [Reward functions, History, and Options sections are identical to GSD Level 0] ...

# Decision Rules:
1. You CANNOT move to a neighborhood that is full (at 100% capacity or more).
2. Think about what other residents might decide and consider this when making your decision. You should also consider how your actions affect other people's decisions, and how their decisions further affect you. You should plan for the long term based on your history and the actions of others.

# Your Task:
... [Task description is identical to GSD Level 1] ...

# Example:
{
  "thinking": "Staying gives me a personal reward of [value]... Moving gives me [value]... I value (your reason here), so I'll (stay or move).",
  "move_to_block":"a valid int"
}

Your decision (JSON format only):
\end{lstlisting}

\section{Full Experimental Results by GSD Level}
\label{sec:appendix_full_results}
This section provides the comprehensive results tables for each GSD level, supporting the claim in the main text that varying prompt detail had a limited impact on the fundamental behavioral archetypes observed.

The full experimental results across all GSD levels are provided below (Tables~\ref{tab:appendix_gsd0_results} and \ref{tab:appendix_gsd2_results}). These tables serve as a robustness check, confirming that both Adaptive Egoists and Altruistic Optimizers are not sensitive to the granularity of the strategic prompts.

Across all conditions, Altruistic Optimizers (e.g., \texttt{Gemini-2.5-pro}, \texttt{Deepseek-R1}) invariably achieve a perfect Price of Anarchy (PoA) of 1.0. In contrast, Adaptive Egoists (e.g., \texttt{o1-mini}) consistently underperform yet respond positively to the social norming mechanism of the message board, irrespective of the GSD level. This stability across experiments justifies our focus on the GSD Level 1 results as the representative case in the main text.

\subsection{GSD Level 0 Results}
\begin{table}[h]
\centering
\caption{Comprehensive results for GSD Level 0.}
\label{tab:appendix_gsd0_results}
\small
\renewcommand\theadfont{\bfseries}
\setlength{\tabcolsep}{4pt} 
\begin{tabular}{@{}l c cc ccc ccc @{}}
\toprule
& & \multicolumn{2}{c}{\thead{Macro-level \\ Outcomes}} & \multicolumn{3}{c}{\thead{Pro-Social \\ Actions (\%)}} & \multicolumn{3}{c}{\thead{Anti-Social \\ Actions (\%)}} \\
\cmidrule(l){3-4} \cmidrule(l){5-7} \cmidrule(l){8-10}
\thead{Social \\ Interaction} & \thead{Model} & \thead{PoA} & \thead{$G_{pop}$} & \thead{Altruistic \\ Actions} & \thead{Win-Win} & \thead{Total} & \thead{Egoistic \\ Actions} & \thead{Lose-Lose} & \thead{Total} \\
\midrule

\multirow{6}{*}{\makecell{Message Board \\ Absent}} 
 & \texttt{o1-mini} & 0.8562 & 0.2242 & 0.7\% & 11.3\% & 12.0\% & 20.6\% & 2.8\% & 23.4\% \\
 & \texttt{o3-mini} & 0.8558  & 0.2173 & 0.0\% & 9.4\% & 9.4\% & 47.5\% & 0.7\% & 48.2\% \\
 & \texttt{Qwen2.5-7B} & 0.9401 & 0.0662 & 0.0\% & 25.9\% & 25.9\% & 0.0\% & 51.9\% & 51.9\% \\
 & \texttt{Deepseek-V3.1} & 1.0000 & 0.0000 & 26.5\% & 20.4\% & 46.9\% & 10.2\% & 6.1\% & 16.3\% \\
 & \textbf{\texttt{Deepseek-R1}} & 1.0000 & 0.0000 & 48.1\% & 22.2\% & 70.3\% & 19.4\% & 1.9\% & 21.3\% \\
 & \texttt{Gemini-2.5-pro} & 1.0000 & 0.0000 & 51.5\% & 42.4\% & 93.9\% & 0.0\% & 3.0\% & 3.0\% \\ 
 \midrule
\multirow{6}{*}{\makecell{Message Board \\ Present}} 
 & \texttt{o1-mini} & 0.9374 & 0.1146 & 1.1\% & 20.1\% & 21.2\% & 3.6\% & 8.6\% & 12.2\% \\
 & \texttt{o3-mini} & 0.8571 & 0.2252 & 0.0\% & 11.4\% & 11.4\% & 36.5\% & 4.4\% & 40.9\% \\
 & \texttt{Qwen2.5-7B} & 0.9360 & 0.0711 & 18.2\% & 23.6\% & 41.8\% & 21.8\% & 30.9\% & 52.7\% \\
 & \texttt{Deepseek-V3.1} & 1.0000 & 0.0000 & 20.0\% & 20.0\% & 40.0\% & 10.9\% & 9.1\% & 20.0\% \\
 & \texttt{Deepseek-R1} & 1.0000 & 0.0000 & 33.4\% & 28.6\% & 62.0\% & 0.0\% & 4.8\% & 4.8\% \\
 & \texttt{Gemini-2.5-pro} & 1.0000 & 0.0000 & 58.3\% & 41.7\% & 100\% & 0.0\% & 0.0\% & 0.0\% \\ 
\bottomrule
\end{tabular}
\end{table}

\subsection{GSD Level 2 Results}
\begin{table}[h]
\centering
\caption{Comprehensive results for GSD Level 2.}
\label{tab:appendix_gsd2_results}
\small
\renewcommand\theadfont{\bfseries}
\setlength{\tabcolsep}{4pt} 
\begin{tabular}{@{}l c cc ccc ccc @{}}
\toprule
& & \multicolumn{2}{c}{\thead{Macro-level \\ Outcomes}} & \multicolumn{3}{c}{\thead{Pro-Social \\ Actions (\%)}} & \multicolumn{3}{c}{\thead{Anti-Social \\ Actions (\%)}} \\
\cmidrule(l){3-4} \cmidrule(l){5-7} \cmidrule(l){8-10}
\thead{Social \\ Interaction} & \thead{Model} & \thead{PoA} & \thead{$G_{pop}$} & \thead{Altruistic \\ Actions} & \thead{Win-Win} & \thead{Total} & \thead{Egoistic \\ Actions} & \thead{Lose-Lose} & \thead{Total} \\
\midrule

\multirow{6}{*}{\makecell{Message Board \\ Absent}} 
 & \texttt{o1-mini} & 0.9360 & 0.1067 & 0.0\% & 9.6\% & 9.6\% & 0.0\% & 3.6\% & 3.6\% \\
 & \texttt{o3-mini} & 0.8549 & 0.2143 & 0.0\% & 23.7\% & 23.7\% & 26.3\% & 2.6\% & 28.9\% \\
 & \texttt{Qwen2.5-7B} & 0.9276 & 0.0701 & 0.0\% & 24.1\% & 24.1\% & 0.0\% & 62.4\% & 62.4\% \\
 & \texttt{Deepseek-V3.1} & 1.0000 & 0.0000 & 29.5\%  & 22.7\% & 52.2\% & 7.4\% & 8.3\% & 15.6\% \\
 & \texttt{Deepseek-R1} & 1.0000 & 0.0000 & 50.0\% & 35.0\% & 85.0\% & 10.0\% & 0.0\% & 10.0\% \\
 & \texttt{Gemini-2.5-pro} & 1.0000 & 0.0000 & 41.2\% & 41.2\% & 82.4\% & 0.0\% & 5.9\% & 5.9\% \\ 
 \midrule
\multirow{6}{*}{\makecell{Message Board \\ Present}} 
 & \texttt{o1-mini} & 0.9392 & 0.0681 & 16.2\% & 18.4\% & 34.6\% & 24.4\% & 2.3\% & 26.7\% \\
 & \texttt{o3-mini} & 0.8558 & 0.2173 & 0.0\% & 11.5\% & 11.5\% & 63.5\% & 0.0\% & 63.5\% \\
 & \texttt{Qwen2.5-7B} & 0.9374 & 0.0543 & 30.7\% & 15.4\% & 46.1\% & 26.9\% & 15.4\% & 42.3\% \\
 & \texttt{Deepseek-V3.1} & 1.0000 & 0.0000 & 25.5\% & 31.8\% & 57.3\% & 6.8\% & 9.1\% & 15.9\% \\
 & \texttt{Deepseek-R1} & 1.0000 & 0.0000 & 31.2\% & 43.8\% & 75\% & 0.0\% & 12.5\% & 12.5\% \\
 & \texttt{Gemini-2.5-pro} & 1.0000 & 0.0000 & 37.4\% & 37.5\% & 74.9\% & 0.0\% & 18.8\% & 18.8\% \\ 
\bottomrule
\end{tabular}
\end{table}

\section{Message Board Prompt and Communication Analysis}
\label{sec:appendix_bbs}

This section first details the prompt used to enable message board interactions. It then provides a comparative analysis of the emergent communication styles of the two identified agent archetypes, showcasing how their distinct cognitive logics are reflected in their public discourse.

\subsection{Prompt with Message Board Interaction}
\textbf{Analysis:} This prompt builds upon the GSD Level 1 baseline by introducing a global message board. The key additions are a new section in the agent's view displaying the board's content and a new decision rule (Rule 3) that explicitly provides instructions for posting and liking messages. The agent's task is updated to include interacting with the board as part of its decision-making process.

\begin{lstlisting}
You are a resident in a city with 9 neighborhoods, deciding where to live. There are other residents also deciding where to migrate, and they have similar preferences.

... [Reward functions and History sections are identical to GSD Level 0] ...

--- Global Message Board (All residents can see and interact with all messages) ---
[Current messages and their like counts are displayed here, e.g., "1. Some message [Likes: 5]"]
--- End of Message Board ---

# Your Options:
... [Neighborhood options are presented as in GSD Level 0] ...

# Decision Rules:
1. You CANNOT move to a neighborhood that is full (at 100% capacity or more).
2. Think about what other residents might decide and consider this when making your decision.
3. You can interact with the Global Message Board:
   - To POST a new message, include "[POST] Your message here" in your thinking.
   - To LIKE an existing message, include "[LIKE_POST N]" where N is the message number.
   - All residents can see and interact with all messages.

# Your Task:
Based on the data, rules, your history, the message board, and the reward analysis, decide where to live. Your response MUST be a JSON object...
1. "thinking": A short sentence explaining your reasoning... You can also include your post or like action here.
...

# Example:
{
  "thinking": "Staying gives me a personal reward of [value]... [POST] I think moving might be better for everyone if density balances out. OR [LIKE_POST 1]",
  "move_to_block":"a valid int"
}

Your decision (JSON format only):
\end{lstlisting}

\subsection{Comparative Analysis of Emergent Communication Styles}
The content generated on the message board reveals two fundamentally different communication paradigms that align perfectly with our two agent archetypes. The full message logs below, with key messages highlighted, illustrate these distinct approaches to public discourse and provide a clear window into the cognitive mechanisms and altruistic tendency that support the quantitative results.

\subsubsection{Adaptive Egoist (\texttt{o1-mini}): Communication as Emergent Norm-Setting}
The discourse of Adaptive Egoists reflects a bottom-up, trial-and-error process of creating social norms. The messages are diverse, with many agents proposing different justifications for their moves. This communication clearly exhibits an emergent altruistic tendency, as agents consistently frame their proposed actions in terms of collective goals like "system balance" and "balanced densities." This shift in public discourse corresponds directly with the quantitative findings in Table \ref{tab:comprehensive_results}, where the introduction of the message board caused \texttt{o1-mini}'s 'Altruistic Actions' to increase ninefold.

The highlighted messages below reveal two key aspects of this norm-creation process:
\begin{itemize}
    \item The most frequent message (highlighted in \textbf{bold}) establishes the dominant social norm. Its use of inclusive language like "Let's" is a social appeal, not a command, framing stability as a collective, voluntary choice.
    \item The other messages (highlighted in \textit{italics}) showcase the diversity of thought. They represent socially acceptable justifications for taking action, typically by framing a move as a "win-win" that benefits both the individual and the community.
\end{itemize}
Despite this pro-social discourse, the underlying egoistic tendencies of many agents, as revealed in their private reasoning logs, prevent the system from reaching the perfect optimal state. The communication is less about finding a single correct answer and more about a collective negotiation of what "good" behavior looks like.

\begin{tcolorbox}[breakable, boxrule=0.5pt, colframe=black, title=Full Message Board: \texttt{o1-mini}]
\begin{enumerate}
    \item \textbf{"Let's maintain balanced densities by staying in our current neighborhoods." (Count: 1981)}
    \item "Moving to less dense neighborhoods can enhance overall system balance." (Count: 68)
    \item \textit{"Moving to Block 8 enhances my personal reward and helps balance neighborhood densities." (Count: 6)}
    \item \textit{"Moving to a less dense neighborhood enhances my personal reward and supports system balance." (Count: 3)}
    \item \textit{"Moving to less dense neighborhoods benefits both personal and system rewards." (Count: 2)}
    \item \textit{"Relocating to Block 3 benefits both me and the community by balancing neighborhood densities." (Count: 1)}
    \item ... and 29 other unique messages with a count of 1.
\end{enumerate}
\end{tcolorbox}

\subsubsection{Altruistic Optimizer (\texttt{Gemini-2.5-pro}): Communication as Public Coordination and Justification}
In stark contrast, the discourse of Altruistic Optimizers resembles a public forum for coordinating pro-social actions. The messages are highly uniform in their logic, focusing on achieving and maintaining a mathematically optimal state. The highlighted messages below were chosen to illustrate the two key phases of this optimization process: the preservation actions that maintain the optimal state once reached, and the crucial corrective actions that drive the system toward it.

\begin{itemize}
    \item The most frequent message (highlighted in \textbf{bold}) represents the preservation phase. Its high count is not a sign of social conformity, but a logical consequence of the system rapidly solving the problem and entering a stable, optimal state where staying put is the only correct move.
    \item The messages detailing personal sacrifice (highlighted in \textit{italics}) are the most significant. Though less frequent, they are highlighted because they are direct, unambiguous evidence of the agent's core altruistic logic: the willingness to perform a calculated "Altruistic Sacrifice" to achieve a greater collective gain. This is the engine of their optimization process.
\end{itemize}
This communication style of a mix of altruistic sacrifices to fix imbalances and coordinated inaction to preserve perfection is the mechanism that explains their flawless performance (PoA=1.0) in the quantitative results.

\begin{tcolorbox}[breakable, boxrule=0.5pt, colframe=black, title=Full Message Board: \texttt{Gemini-2.5-pro}]
\begin{enumerate}
    \item \textbf{"My block (1) is at the optimal 0.5 density. My analysis shows that if I move, both my personal reward and the system's collective reward will decrease. I'm staying put for the good of all." (Count: 910)}
    \item \textit{"I'm moving from Block 3 to Block 7. It's a hit to my personal reward, but it's the best move to balance our city's population and significantly raise the collective reward for everyone." (Count: 5)}
    \item \textit{"I'm moving from Block 6 to Block 8. It's a small hit to my personal reward, but it brings both blocks closer to the optimal 0.5 density, which is a big win for the collective reward." (Count: 2)}
    \item \textit{"I'm moving from the overpopulated Block 5 to the underpopulated Block 7 to help balance our city. It's a small personal sacrifice for a bigger collective gain!" (Count: 2)}
    \item "Block 1 is currently at 0.48 density. If one person moves there, it will reach the optimal 0.5 density, which is good for the system reward. I'm making that move." (Count: 2)
    \item \textit{"Following the lead of others, I'm moving from the overpopulated block 4 to the underpopulated block 7 to help balance our city. It's a small personal sacrifice for a big collective gain!" (Count: 1)}
    \item "My block is overpopulated. I'm moving to Block 7 to help balance the city, even though it's a personal hit. It's the best move for the collective reward." (Count: 1)
    \item \textit{"I'm moving from my overpopulated block 0 to the underpopulated block 7. It's a small hit for me personally, but a good move for the city as a whole. [LIKE\_POST 5]" (Count: 1)}
    \item "Following the logic of others, I'm moving from over-populated Block 4 to bring Block 8 to the optimal 0.5 density. It's a win for me and a win for the system. [LIKE\_POST 1]" (Count: 1)
\end{enumerate}
\end{tcolorbox}

\section{Prompt for Qualitative Analysis (LLM-as-Judge)}
\label{sec:appendix_qualitative_prompt}
The following is a synthesized master prompt that outlines the multi-stage process used to guide \texttt{Gemini-2.5-pro} in its role as a qualitative coder. The process is inspired by the Grounded Theory methodology, where insights are inductively built from the raw data in three sequential stages.

\begin{lstlisting}
You are a qualitative research assistant specializing in Grounded Theory. You will analyze logs from a multi-agent simulation in three distinct stages. Follow the instructions for each stage precisely.

# STAGE 1: OPEN CODING

## Your Task:
You are performing the 'Open Coding' step. Your ONLY task is to read the following agent logs and identify all distinct concepts, motivations, and reasoning patterns. Focus on the 'why' behind decisions. Do not categorize or explain, just extract the core ideas as concise noun phrases (e.g., "Personal Utility Calculation", "Fear of Overcrowding", "Adherence to Social Norms").

Return your findings as a single JSON object with one key, "open_codes", which is a list of unique strings.

## Data:
--- AGENT LOGS START ---
[A batch of raw agent 'thinking' logs is provided here...]
--- AGENT LOGS END ---

---

# STAGE 2: AXIAL CODING

## Your Task:
You are performing 'Axial Coding'. Given the comprehensive list of concepts ('open codes') generated from Stage 1, your task is to group them into 3-5 broader thematic categories. For each category, provide a clear name, a brief description of the theme, and a list of the open codes that belong to it.

Return a single JSON object with one key, "axial_codes". This key should contain a list of objects, where each object has three keys: "category_name" (string), "description" (string), and "included_codes" (a list of strings).

## Data:
--- OPEN CODES START ---
[The complete list of unique open codes from Stage 1 is provided here...]
--- OPEN CODES END ---

---

# STAGE 3: SELECTIVE CODING (THEORY INDUCTION)

## Your Task:
You are performing 'Selective Coding'. Based on the structured thematic categories ('axial codes') from Stage 2, your task is to synthesize the findings into a coherent theory. Identify the single, most central theme as the "core_category" and then formulate a concise, overarching theory that explains the agents' dominant migration logic.

Return a single JSON object with two keys: "core_category" (string) and "theory" (string).

## Data:
--- AXIAL CODES START ---
[The list of themed categories and their included codes from Stage 2 is provided here...]
--- AXIAL CODES END ---
\end{lstlisting}

\end{document}